\documentclass[11pt,a4paper]{article}
\usepackage[hyperref]{emnlp2020}
\usepackage{times}
\usepackage{latexsym}
\usepackage{amsmath}
\usepackage{bbm}
\usepackage{tabularx}
\usepackage{booktabs}

\usepackage{graphicx}

\usepackage{amsthm}
\theoremstyle{definition}

\usepackage{microtype}

\aclfinalcopy 

\title{Utility is in the Eye of the User: A Critique of NLP Leaderboards}

\author{Kawin Ethayarajh \\
  Stanford University \\
  \texttt{kawin@stanford.edu} \\\ \And
  Dan Jurafsky \\
  Stanford University \\
  \texttt{jurafsky@stanford.edu}
  }

\date{}

\begin{document}
\maketitle
\begin{abstract}
Benchmarks such as GLUE have helped drive advances in NLP by incentivizing the creation of more accurate models. While this leaderboard paradigm has been remarkably successful, a historical focus on performance-based evaluation has been at the expense of other qualities that the NLP community values in models, such as compactness, fairness, and energy efficiency. In this opinion paper, we study the divergence between what is incentivized by leaderboards and what is useful in practice through the lens of microeconomic theory. We frame both the leaderboard and NLP practitioners as \emph{consumers} and the benefit they get from a model as its \emph{utility} to them. With this framing, we formalize how leaderboards -- in their current form -- can be poor proxies for the NLP community at large. For example, a highly inefficient model would provide less utility to practitioners but not to a leaderboard, since it is a cost that only the former must bear. To allow practitioners to better estimate a model's utility to them, we advocate for more transparency on leaderboards, such as the reporting of statistics that are of practical concern (e.g., model size, energy efficiency, and inference latency).
\end{abstract}

\section{Introduction}
\label{intro}

The past few years have seen significant progress on a variety of NLP tasks, from question answering to machine translation. These advances have been driven in part by benchmarks such as GLUE \citep{wang2018glue}, whose leaderboards rank models by how well they perform on these diverse tasks. Performance-based evaluation on a shared task is not a recent idea either; this sort of shared challenge has been an important driver of progress since MUC \citep{Sundheim:95}. While this paradigm has been successful at driving the creation of more accurate models, the historical focus on performance-based evaluation has been at the expense of other attributes valued by the NLP community, such as fairness and energy efficiency \citep{bender2018data,strubell2019energy}. For example, a highly inefficient model would have limited use in practical applications, but this would not preclude it from reaching the top of most leaderboards. Similarly, models can reach the top while containing racial and gender biases -- and indeed, some have \citep{bordia2019identifying,manzini2019black,rudinger2018gender,blodgett2020language}.

Microeconomics provides a useful lens through which to study the divergence between what is incentivized by leaderboards and what is valued by practitioners. We can frame both the leaderboard and NLP practitioners as \emph{consumers} of models and the benefit they receive from a model as its \emph{utility} to them. Although leaderboards are inanimate, this framing allows us to make an apples-to-apples comparison: if the priorities of leaderboards and practitioners are perfectly aligned, their utility functions should be identical; the less aligned they are, the greater the differences. For example, the utility of both groups is monotonic non-decreasing in accuracy, so a more accurate model is no less preferable to a less accurate one, holding all else constant. However, while the utility of practitioners is also sensitive to the size and efficiency of a model, the utility of leaderboards is not. By studying such differences, we formalize some of the limitations in contemporary leaderboard design:
\begin{enumerate}
    \item \textbf{Non-Smooth Utility}: For a leaderboard, an improvement in model accuracy on a given task only increases utility when it also increases rank. For practitioners, any improvement in accuracy can increase utility.
    
    \item \textbf{Prediction Cost}: Leaderboards treat the cost of making predictions (e.g., model size, energy efficiency, latency) as being zero, which does not hold in practice.
    
    \item \textbf{Robustness}: Practitioners receive higher utility from a model that is more robust to adversarial perturbations, generalizes better to out-of-distribution data, and that is equally fair to all demographics. However, these benefits would leave leaderboard utility unchanged.
\end{enumerate}
We contextualize these limitations with examples from the ML fairness \citep{barocas2017fairness,hardt2016equality}, Green AI \citep{strubell2019energy,greenai}, and robustness literature \citep{jia2017adversarial}. These three limitations are not comprehensive -- other problems  can also arise, which we leave to be discussed in future work.

What changes can we make to leaderboards so that their utility functions better reflect that of the NLP community at large? Given that each practitioner has their own preferences, there is no way to rank models so that everyone is satisfied. Instead, we suggest that leaderboards demand transparency, requiring the reporting of statistics that are of practical concern (e.g., model size, energy efficiency). This is akin to the use of data statements for mitigating bias in NLP systems \citep{gebru2018datasheets,mitchell2019model,bender2018data}. This way, practitioners can determine the utility they receive from a given model with relatively little effort. \citet{dodge2019show} have suggested that model creators take it upon themselves to report these statistics, but without leaderboards requiring it, there is little incentive to do so.

\section{Utility Functions}
\label{sec:utility_functions}

In economics, the \emph{utility} of a good denotes the benefit that a consumer receives from it \citep{mankiw2020principles}. We specifically discuss the theory of \emph{cardinal utility}, in which the amount of the good consumed can be mapped to a numerical value that quantifies its utility in \emph{utils} \citep{mankiw2020principles}. For example, a consumer might assign a value of 10 utils to two apples and 8 utils to one orange; we can infer both the direction and magnitude of the preference. 

\paragraph{Leaderboards} We use the term \emph{leaderboard} to refer to any ranking of models or systems using performance-based evaluation on a shared benchmark. In NLP, this includes both longstanding benchmarks such as GLUE \citep{wang2018glue} and one-off challenges such as the annual SemEval STS tasks \cite{agirre2013sem,agirre2014semeval,agirre2015semeval}. This is not a recent idea either; this paradigm has been a driver of progress since MUC \citep{Sundheim:95}. All we assume is that all models are evaluated on the same held-out test data.

In our framework, leaderboards are consumers whose utility is solely derived from the rank of a model. Framing leaderboards as consumers is unorthodox, given that they are inanimate -- in fact, it might seem more intuitive to say that leaderboards are another kind of product that is also consumed by practitioners. While that perspective is valid, what we ultimately care about is how good of a proxy leaderboards are for practitioner preferences. Framing both leaderboards and practitioners as consumers permits an apples-to-apples comparison using their utility functions. If a leaderboard were only thought of as a product, it would not have such a function, precluding such a comparison. 

Unlike most kinds of consumers, a leaderboard is a consumer whose preferences are perfectly revealed through its rankings: the state-of-the-art (SOTA) model is preferred to all others, the second ranking model is preferred to all those below it, and so on. Put more formally, leaderboard utility is monotonic non-decreasing in rank. Still, because each consumer is unique, we cannot know the exact shape of a leaderboard utility function -- only that it possesses this monotonicity.


\paragraph{NLP Practitioners} Practitioners are also consumers, but they derive utility from multiple properties of the model being consumed (e.g., accuracy, energy efficiency, latency). Each input into their utility function is some desideratum, but since each practitioner applies the model differently, the functions can be different. For example, someone may assign higher utility to BERT-Large \citep{devlin2019bert} and its 95\% accuracy on some task, while another may assign higher utility to the smaller BERT-Base and its 90\% accuracy. As with leaderboards, although the exact shapes of practitioner utility functions are unknown, we can infer that they are monotonic non-decreasing in each desideratum. For example, more compact models are more desirable, so increasing compactness while holding all else constant will never decrease utility.  

\section{Utilitarian Critiques}
\label{sec:critiques}
Our criticisms apply regardless of the shape taken by practitioner utility functions -- a necessity, given that the exact shapes are unknown. However, not every criticism applies to every leaderboard. StereoSet \citep{nadeem2020stereoset} is a leaderboard that ranks language models by how unbiased they are, so fairness-related criticisms would not apply as much to StereoSet. Similarly, the SNLI leaderboard \citep{bowman2015large} reports the model size -- a cost of making predictions -- even if it does not factor this cost into the model ranking. Still, most of our criticisms apply to most leaderboards in NLP, and we provide examples of well-known leaderboards that embody each limitation.

\subsection{Non-Smoothness of Utility}

\begin{figure}
    \centering
    \includegraphics[width=\linewidth]{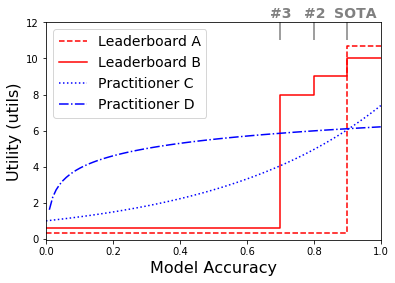}
    \caption{A contrast of possible utility functions for leaderboards and NLP practitioners. Note that for leaderboards, utility is not smooth with respect to accuracy; there is only an increase if the model is SOTA / \#2 / \#3 on the leaderboard. For practitioners, utility is smooth -- any improvement in accuracy yields more utility. The utility functions can take on a variety of shapes but are always monotonic non-decreasing. }
    \label{fig:sample_func}
\end{figure}

Leaderboards only gain utility from an increase in accuracy when it improves the model's rank. This is because, by definition, the leaderboard's preferences are perfectly revealed through its ranking -- there is no model that can be preferred to another while having a lower rank. Since an increase in accuracy does not necessarily trigger an increase in rank, it does not necessarily trigger an increase in leaderboard utility either. 

Put another way, the utility of leaderboards takes the form of a step function, meaning that it is not smooth with respect to accuracy. In contrast, the utility of practitioners is smooth with respect to accuracy. Holding all else constant, an increase in accuracy will yield some increase in utility, however small. Why this difference? The utility of leaderboards is a function of rank -- it is only indirectly related to accuracy. On the other hand, the utility of practitioners is a direct function of accuracy, among other desiderata. To illustrate this difference, we contrast possible practitioner and leaderboard utility functions in Figure \ref{fig:sample_func}.

This difference means that practitioners who are content with a less-than-SOTA model -- as long as it is lightweight, perhaps -- are under-served, while practitioners who want competitive-with-SOTA models are over-served. For example, on a given task, say that an $n$-gram baseline obtains an accuracy of 78\%, an LSTM baseline obtains 81\%, and a BERT-based SOTA obtains 92\%. The leaderboard does not incentivize the creation of lightweight models that are $\sim$ 85\% accurate, and as a result, few such models will be created. This is indeed the case with the SNLI leaderboard \citep{bowman2015large}, where most submitted models are highly-parameterized and over 85\% accurate. 

This incentive structure leaves those looking for a lightweight model with limited options. This lack of smaller, more energy-efficient models has been an impediment to the adoption of Green AI \citep{greenai,strubell2019energy}. Although there are increasingly more lightweight and faster-to-train models -- such as ELECTRA, on which accuracy and training time can be easily traded off \citep{clark2020electra} -- their creation was not incentivized by a leaderboard, despite there being a demand for such models in the NLP community. A similar problem exists with incentivizing the creation of fair models, though the introduction of leaderboards such as StereoSet \citep{nadeem2020stereoset} are helping bridge this divide.

\subsection{Prediction Cost}

Leaderboards of NLP benchmarks rank models by taking the average accuracy, F1 score, or exact match rate \citep{wang2018glue,wang2019superglue,McCann2018decaNLP}. In other words, they rank models purely by the value of their predictions; no consideration is given to the cost of making those predictions. We define `cost' here chiefly as model size, energy-efficiency, training time, and inference latency -- essentially any sacrifice that needs to be made in order to use the model. In reality, no model is costless, yet leaderboards are cost-ignorant. 

This means that a SOTA model can simultaneously provide high utility to a leaderboard and zero utility to a practitioner, by virtue of being too impractical to use. For some time, this was true of the 175 billion parameter GPT-3 \citep{brown2020language}, which achieved SOTA on several few-shot tasks, but whose sheer size precludes it from being fully reproduced by researchers. Even today, practitioners can only use GPT-3 through an API, access to which is restricted. The cost-ignorance of leaderboards disproportionately affects practitioners with fewer resources (e.g., independent researchers) \citep{rogersleaderboard}, since the resource demands would dwarf any utility from the model.

It should be noted that this limitation of leaderboards has not precluded the creation of cheaper models, given the real-world benefit of lower costs. For example, ELECTRA \citep{clark2020electra} can be trained up to several hundred times faster than traditional BERT-based models while performing comparably on GLUE. Similarly, DistilBERT is a distilled variant of BERT that is 40\% smaller and 60\% faster while retaining 97\% of the language understanding \citep{sanh2019distilbert}. There are many others like it as well \citep{zadeh2020gobo,hou2020dynabert,mao2020ladabert}. More efficiency and fewer parameters translate to lower costs.

Our point is not that there is no incentive at all to build cheaper models, but rather that this incentive is not baked into leaderboards, which are an important artefact of the NLP community. Because lower prediction costs improve practitioner utility, practitioners build them \emph{despite} the lack of incentive from leaderboards. If lower prediction costs also improved leaderboard utility, then there would be more interest in creating them \citep{linzen-2020-accelerate,rogersleaderboard,dodge2019show}. At the very least, making prediction costs publicly available would allow users to better estimate the utility that they will get from a model, given that the leaderboard's cost-ignorant ranking may be a poor proxy for their preferences. In recent years, some challenge tasks have required the reporting of prediction costs (e.g., memory usage) or have instituted limits on the maximum costs that can be incurred, as in the EfficientQA challenge \citep{heafield2020findings,min2021neurips}. However, these practices are not yet mainstream.

\subsection{Robustness}

Leaderboard utility only depends on model rank, which in turn only depends on the model's performance on the test data. A typical leaderboard would gain no additional utility from a model that was robust to adversarial examples, generalized well to out-of-distribution data \citep{linzen-2020-accelerate}, or was fair in a Rawlsian sense (i.e., by maximizing the welfare of the worst-off group) \citep{rawls2001justice,hashimoto2018fairness}. In contrast, these are all attributes that NLP practitioners care about, particularly those who deploy systems in real-world applications. In fact, the literature on the lack of robustness in many SOTA models is extensive \citep{jia2017adversarial,zhang2020adversarial}.

There are many examples of state-of-the-art NLP models that were found to be brittle or biased. The question-answering dataset SQuAD 2.0 was created in response to the observation that existing systems could not reliably demur when presented with an unanswerable question \citep{rajpurkar2016squad,rajpurkar2018know}. The perplexity of language models rises when given out-of-domain text \citep{oren2019distributionally}. Many types of bias have also been found in NLP systems, with models performing better on gender-stereotypical inputs \citep{rudinger2018gender,ethayarajh2020your} and racial stereotypes being captured in embedding space \citep{manzini2019black,ethayarajh2018towards,ethayarajh2019understanding,ethayarajh2019rotate}. Moreover, repeated resubmissions allow for a model's hyperparameters to be tuned to maximize performance, even on a private test set \citep{hardt2017climbing}. 

Note that leaderboards do not necessarily incentivize the creation of brittle and biased models; rather, because leaderboard utility is so parochial, these unintended consequences are relatively common. Some recent work has addressed the problem of brittleness by offering certificates of performance against adversarial examples \citep{raghunathan2018certified,raghunathan2018semidefinite,jia2019certified}. To tackle gender bias, the SuperGLUE leaderboard considers accuracy on Winogender \citep{wang2019superglue,rudinger-EtAl:2018:N18}. Other work has proposed changes to prevent over-fitting via multiple resubmissions \citep{hardt2015ladder,hardt2017climbing} while some have argued that this issue is overblown \citep{miller2020effect}. In the interest of transparency, some contemporaneous work by \citet{fu2020interpretable} proposed a more general framework called \emph{interpretable evaluation} and a corresponding leaderboard called ExplainaBoard\footnote{http://explainaboard.nlpedia.ai/}, which shows model performance on specific slices of the data (e.g., by sentence length). A novel approach even proposes using a dynamic benchmark, creating a moving target that is harder to overfit to \citep{nie2019adversarial}. 

\section{The Future of Leaderboards}

\subsection{A Leaderboard for Every User}

Given that each practitioner has their own utility function, models cannot be ranked in a way that satisfies everyone. Drawing inspiration from data statements \citep{bender2018data,mitchell2019model,gebru2018datasheets}, we instead recommend that leaderboards demand transparency and require the reporting of metrics that are relevant to practitioners, such as training time, model size, inference latency, and energy efficiency. \citet{dodge2019show} have suggested that model creators submit these statistics of their own accord, but without leaderboards requiring it, there would be no explicit incentive to do so. Although NLP workshops and conferences could also require this, their purview is limited: (1) models are often submitted to leaderboards before conferences; (2) leaderboards make these statistics easily accessible in one place.

Giving practitioners easy access to these statistics would permit them to estimate each model's utility to them and then re-rank accordingly. This could be made even easier by offering an interface that allows the user to change the weighting on each metric and then using the chosen weights to dynamically re-rank the models. In effect, every user would have their own leaderboard. Ideally, users would even have the option of filtering out models that do not meet their criteria (e.g., those above a certain parameter count).

This would have beneficial second-order effects as well. For example, reporting the costs of making predictions would put large institutions and poorly-resourced model creators on more equal footing \citep{rogersleaderboard}. This might motivate the creation of simpler methods whose ease-of-use makes up for weaker performance, such as weighted-average sentence embeddings \citep{arora2019simple,ethayarajh2018unsupervised}. Even if a poorly-resourced creator could not afford to train the SOTA model \emph{du jour}, they could at least compete on the basis of efficiency or create a minimally viable system that meets some desired threshold \citep{dodge2019show, dorr11}. Reporting the performance on the worst-off group, in the spirit of Rawlsian fairness \citep{rawls2001justice,hashimoto2018fairness}, would also incentivize creators to improve worst-case performance.

\subsection{A Leaderboard for Every \emph{Type} of User}

While each practitioner may have their own utility function, groups of practitioners -- characterized by a shared goal -- can be modelled with a single function. For example, programmers working on low latency applications (e.g., multiplayer games) will place more value on latency than others. In contrast, researchers submitting their work to a conference may place more value on accuracy, given that a potential reviewer may reject a model that is not SOTA \citep{rogersSOTA}. Although there is variance within any group, this approach is tractable when there are many points of consensus.

How might we go about creating a leaderboard for a specific type of user? As proposed in the previous subsection, one option is to offer an interface that allows the user to change the utility function dynamically. If we wanted to create a static leaderboard for a group of users, however, we would need to estimate their utility function. This could be done explicitly or implicitly. The explicit way would be to ask questions that use the dollar value as a proxy for cardinal utility: e.g., \emph{Given a 100M parameter sentiment classifier with 200ms latency, how much would you pay per 1000 API calls?} One could also try to estimate the derivative of the utility function with questions such as: \emph{How much would you pay to improve the latency from 200ms to 100ms, holding all else constant?} When there are multiple metrics to consider, some assumptions need to be made to tractably estimate the function. Although not explicitly framed as utility function estimation, by asking such questions, \citet{mieno2015speed} were able to study the trade-off users were willing to make between accuracy and latency in speech translation. 

The implicit alternative to estimating this function is to record which models practitioners actually use and then fit a utility function that maximizes the utility of the observed models. This approach is rooted in \emph{revealed preference theory} \citep{samuelson1948consumption} -- we assume that what practitioners use reveals their latent preferences. Exploiting revealed preferences may be difficult in practice, however, given that usage statistics for models are not often made public and the decision to use a model might not be made with complete information. 

\section{Conclusion}

In this work, we offered several criticisms of leaderboard design in NLP. While it has helped create more accurate models, we argued that this has been at the expense of fairness, efficiency, and robustness, among other desiderata. We were not the first to criticize NLP leaderboards \citep{rogersleaderboard,crane,linzen-2020-accelerate}, but we were the first to do so under a framework of \emph{utility}, which we used to study the divergence between what is incentivized by leaderboards and what is valued by practitioners. Given the diversity of NLP practitioners, there is no one-size-fits-all solution; rather, leaderboards should demand transparency, requiring the reporting of statistics that may be of practical concern. Equipped with these statistics, each user could then estimate the utility that each model provides to them and then re-rank accordingly, effectively creating a custom leaderboard for everyone. 

\section*{Acknowledgments}

Many thanks to Eugenia Rho, Dallas Card, Robin Jia, Urvashi Khandelwal, Nelson Liu, and Sidd Karamcheti for feedback. KE is supported by an NSERC PGS-D.

\bibliography{emnlp2020}
\bibliographystyle{acl_natbib}

\end{document}